\documentclass[runningheads]{llncs}
\usepackage[T1]{fontenc}
\usepackage{graphicx}
\usepackage{hyperref}
\usepackage{color}

\usepackage{booktabs}
\usepackage{multirow}
\usepackage{array}
\usepackage{rotating}
\usepackage{adjustbox}
\usepackage{subcaption}

\usepackage{siunitx}
\usepackage{amsmath}
\usepackage{amssymb}

\usepackage{xcolor}

\begin{document}
\title{How Swarms Differ: Challenges in Collective Behaviour Comparison}
\author{André Fialho Jesus\inst{1, 2}\orcidID{0009-0001-7490-2133} \and
Jonas Kuckling\inst{1, 2, 3}\orcidID{0000-0003-2391-2275}}
\authorrunning{A. Fialho Jesus and J. Kuckling}
\institute{Department of Computer and Information
Science, University of Konstanz, Germany\\ \and 
Centre for the Advanced Study of Collective Behaviour, University of Konstanz, Germany \and
Zukunftskolleg, University of Konstanz, Germany.
\email{\{andre.jesus,jonas.kuckling\}@uni-konstanz.de}
}
\index{Fialho Jesus, André}
\index{Kuckling, Jonas}
\maketitle              % typeset the header of the contribution
\begin{abstract}
Collective behaviours often need to be expressed through numerical features, e.g., for classification or imitation learning.
This problem is often addressed by proposing an ad-hoc feature set for a particular swarm behaviour context, usually without further consideration of the solution's resilience outside of the conceived context.
Yet, the development of automatic methods to design swarm behaviours is dependent on the ability to measure quantitatively the similarity of swarm behaviours.
Hence, we investigate the impact of feature sets for collective behaviours.
We select swarm feature sets and similarity measures from prior swarm robotics works, which mainly considered a narrow behavioural context and assess their robustness.  
We demonstrate that the interplay of feature set and similarity measure makes some combinations more suitable to distinguish groups of similar behaviours.
We also propose a self-organised map-based approach to identify regions of the feature space where behaviours cannot be easily distinguished.

% \keywords{Features of collective behaviour \and Behavioural similarity.}
\end{abstract}
%
%
%
%%%%%%%%%%%%%%%%%%%%%%%%%%%%%%%%%%%%%%%%%%%%%%%%%%%%%%%%%%%%%%%%%%%%%%%%%%%%%%%%
\section{Introduction}
\label{sec:introduction}
% >>>>> Introduce the problem
% ...

%% ABT structure
A central problem of automatic design methods for swarm robotics applications is how to measure the similarity between different collective behaviours.
In particular, the similarity between collective behaviours has been explored in the context of novelty search~\cite{GomChr2013gecco}, classification~\cite{YanSamAru-etal2023IEEEACCESS} and imitation learning~\cite{AlhAbdHau2022ants,GhaKucGarBir2023icra}.
% Furthermore, one could even consider the reward or fitness function as a particular (if rudimentary) way of specifying behavioural similarity (i.e., it assumes that maximising this objective function will yield a behaviour similar to the desired one~\cite{BirLigBoz-etal2019FRAI}).
% \section{Related Work}
% \label{sec:related-work}
% >>>>> What is the prior work? How is your method different from prior work?
% Swarm features and similarity measures have been studied in different contexts.
Yang et al.~\cite {YanSamAru-etal2023IEEEACCESS} propose a set of swarm features and use several machine learning methods to recognise collective behaviours, including flocking and random walk.
%Navarro and Matía~\cite{NavMat2009gem} propose swarm features to describe the shape, movement, algorithmic quality, and efficiency, to quantify collective motion behaviours.
%Similarly, Harwell and Gini~\cite{HarGin2019ijcai} propose a feature set and test it with swarms of up to $\num{10000}$ robots, which was later extended and applied to two swarm robotics tasks~\cite{HarGin2020arxiv}.
Gomes and Christensen~\cite{GomChr2013gecco} provide two measures of swarm behaviour similarity.
% Features of collective behaviour are also often considered in the context of imitation learning.
Alharthi et al.~\cite{AlhAbdHau2022ants} provide a set of swarm features to an evolutionary algorithm tasked to correctly reconstruct the behaviour tree that simulated the original collective behaviour.
The same authors extend the previous approach by proposing a new set of environmental imprint features~\cite{AlhAbdHau2025eurogp}.
% In inverse reinforcement learning, a reward function is derived from the available features.
Gharbi et al.~\cite{GhaKucGarBir2023icra} proposed a swarm feature set for inverse reinforcement learning that was reused in a similar context in a follow-up paper by Szpirer et al.~\cite{SzpGarBir2024iros}.
% Rahman et al.~\cite{RahWhiCro-etal2023acc} define several individual and socialpolicies and learn a linear combination of them that best describes the collective behaviours.

While humans can recognise different collective behaviours with relative ease~\cite{YanSamAru-etal2023IEEEACCESS}, automatic methods need to rely on numerical features describing the collective behaviour (e.g.,~\cite{AlhAbdHau2022ants,GhaKucGarBir2023icra,YanSamAru-etal2023IEEEACCESS}).
Usually, the feature sets are developed in an ad-hoc fashion, targeting a particular behavioural context.
As a result, it is unclear if these feature sets hold beyond the context in which they are proposed.
This limits their usability in fully automatic design~\cite{BirLigHas2020NATUMINT}.
Consequently, we explore different ad-hoc feature sets and their implications in comparing collective behaviours.
In particular, we focus on computing similarity (or distance) measures and classification.
Similarity measures provide us with an immediate numerical value that can be used in fully automatic methods.
Classification addresses a slightly different aspect of behavioural similarity.
By focusing on the differences between behaviour classes, it highlights the feature set's discriminative power.
%Similarly, by identifying regions in which behaviours are typically misclassified, we can gain insights into behavioural aspects not adequately covered by the feature set.

\section{Methodology}
\label{sec:method}
% >>>>> How does your method work and why should it work

% We build a dataset with six collective behaviours and extract features  to characterise the state of the swarm over time.
% Then  we investigate how the swarm feature sets impact the behavioural similarity  and how the features contribute to the behaviours identification.

\paragraph{Collective behaviours}
\label{subsec:swarm-behaviours}

%In order to assess the impact of feature sets on similarity and classification assessments, we build a dataset of six collective behaviours that are typically associated with spatial organisation and coordinated motion~\cite{SchUmlSenElm2020FRAI}.
We consider six collective behaviours that are typically associated with spatial organisation and coordinated motion~\cite{SchUmlSenElm2020FRAI}.
The behaviours are executed on a swarm of massless points in an otherwise empty environment.
We control the initial placement of the agents and subsequently the execution of the (deterministic) behaviours through a random seed. 
The agents' speeds are kept constant to facilitate the comparison of behaviours.
We consider two variants of the flocking coordinated motion: the \textit{Reynolds} boids model~\cite{Rey1987siggraph} and the \textit{Vicsek} model~\cite{VicCziBen-etal1995PRL}.
For spatial organisation behaviours, we consider \textit{aggregation} (inspired by Kaiser et al.~\cite{KaiBegPla-etal2022icra}), \textit{dispersion} (inspired by Kaiser et al.~\cite{KaiBegPla-etal2022icra}), and two variants of random walk: \textit{ballistic motion} and \textit{Brownian motion} (following Kegeleirs et al.~\cite{KegGarBir2019taros}).
See the supplementary video for the example trajectories of the collective behaviours.
In terms of similarity, we expect that the flocking behaviours are identified as highly similar to each other.
\textit{Aggregation} should have a moderate similarity degree to either flocking behaviour, and the two random walks and \textit{dispersion} should, for the most part, be dissimilar to them.

\paragraph{Feature sets}
\label{subsec:swarm_char}

\begin{table}[tb]
\centering
\caption{Feature sets considered in this work.}
\label{tab:feature_encodings}
\renewcommand{\arraystretch}{1.25}
\setlength{\tabcolsep}{6pt}
\begin{tabular}{@{}
>{\raggedright\arraybackslash}p{0.96\textwidth}
@{}}
\toprule

\textsc{Alharthi2022}~\cite{AlhAbdHau2022ants}: 
maximum swarm shift \textbullet\ 
center of mass \textbullet\ 
swarm mode index \textbullet\ 
longest path \textbullet\ 
maximum radius \textbullet\ 
average local density \textbullet\ 
average nearest neighbour distance \textbullet\ 
beta index.\;
\textsc{Gomes2013}~\cite{GomChr2013gecco}: 
boid x \textbullet\ 
boid y \textbullet\ 
boid vx \textbullet\ 
boid vy.\;
\textsc{Yang2023}~\cite{YanSamAru-etal2023IEEEACCESS}: 
collision count \textbullet\ 
flock density \textbullet\ 
grouping \textbullet\ 
straggler count \textbullet\ 
order \textbullet\ 
subgroup count.\;
\textsc{Gharbi2023}~\cite{GhaKucGarBir2023icra}: 
neighbour shortest distances.\\

\bottomrule
\end{tabular}
\end{table}

We utilise the previously defined collective behaviours to compute and extract four feature sets from the literature (see Table~\ref{tab:feature_encodings}).
We hypothesise that different feature sets yield measurable trade-offs in the similarity assessment and classification.
For the full details of the features, please consult the original sources.
``Alharthi2022''~\cite{AlhAbdHau2022ants,AlhAbdHau2025eurogp} offers a comprehensive set of eight swarm-level features and has been previously used to assess the similarity of collective behaviours.
``Gomes2013''~\cite{GomChr2013gecco} offers a set of agent-level features, based on their sensor and actuator information, which has been used to derive generic measures of collective behaviour.
``Yang2023''~\cite{YanSamAru-etal2023IEEEACCESS} comprises six swarm-level features, which have been used to classify collective behaviours of boid-like particles.
Note that in ``Yang2023'', we opted to discard the diffusion feature as we were unable to find a definitive source on how to compute it.
Lastly, we consider ``Gharbi2023''~\cite{GhaKucGarBir2023icra,SzpGarBir2024iros}, a set of environment-based agent-level features. % previously used in an inverse reinforcement learning context.
We acknowledge that the construction of this feature set degenerates to only inter-robot distances in our environment.
We opted, however, to include this feature set as a lower baseline to put the performance of other feature sets into context.

\paragraph{Similarity Assessment}
\label{subsec:swarm_simil}
We consider the impact of the swarm features on four similarity measurements: cosine similarity, Euclidean distance, combined state count~\cite{GomChr2013gecco}, and sampled average state~\cite{GomChr2013gecco}.
\emph{Cosine similarity} is a standard measure to quantify similarity using vector representations.
Due to the construction of our features, the cosine similarity scores are between $[0, 1]$, where $1$ refers to perfectly equal features.
Likewise, the \emph{Euclidean distance} is another standard measure of similarity.
As a distance measure, two identical feature vectors will have a distance of $0$.
The \emph{combined state count} method  (see~\cite{GomChr2013gecco}) discretises the swarm feature set (describing the state space) and counts the number of times any swarm member is in a given state.
When the behaviours are identical, it leads to a score of $0$ and $1$ when they do not have anything in common.
The \emph{sampled average state} method (see~\cite{GomChr2013gecco}) summarises each feature by averaging its value over all robots and splitting it over several time windows.
Since it is a distance measure, the closer to $0$, the more similar the behaviours are.

\paragraph{Explainable Classification}
\label{subsec:swarm_classif}
The similarity and distance measures do not provide insights towards which features are important to distinguish behaviours or why certain behaviours score high similarity while being visually distinct.
Therefore, we consider a explainable classification algorithm to gain further insights into these questions.
In particular, we employ a self-organising map (SOM) for classification.
Self-organising maps~\cite{Koh1982BC,Koh1990NEUROCOMP} are a type of prototype-based classifiers trained through the competitive learning paradigm~\cite{RumZip1985COGSCI}.
% In competitive learning, the network nodes compete, in an unsupervised way, to represent subsets of the data.
% This provides two main benefits~\cite{GamelasSousa2015RobustClassificationReject}.
% Firstly, the local explanatory rules associated with each prototype facilitate the interpretability of the decision-making.
% Secondly, prototypes can be easily added or deleted to approximate evolving and/or non-stationary data distributions.
The map topology allows us to define an inter-node distance between units.
During the training phase, the \emph{best matching unit} for each training sample is selected.
The weights associated with the best-matching unit and its neighbours are adjusted towards the training sample.
After training, the nodes' classification labels are extracted according to the ``majority voting'' strategy~\cite{CoeBaaMed2017wsom}.
In case of a tie for the node label choice, one of the tied labels is selected randomly, and when a node has no samples assigned to it, the most commonly assigned class label amongst the SOM nodes is used as the default class.
% Besides the classification accuracy, we report on the quantisation and topological SOM errors.
% The quantisation error captures the average Euclidean distance error made in the data projection in the SOM, whereas the topological error defines the fraction of the data samples whose best and second-best matching units are not neighbours on the map~\cite{forest2020surveyimplementationperformancemetrics}.

\section{Experimental Setup}
\label{sec:experimental_setup}
% >>>>> ...

All behaviours are executed in a $500\text{px}\times500\text{px}$ environment that can either be bounded (agents cannot pass beyond the boundaries) or unbounded (agents that pass beyond a boundary appear on the other side).
% The behaviours are simulated for $1500$ steps with an update rate of $\SI{10}{\hertz}$.
% The agents have a sensing range of $100$ pixels, in which they can detect other agents and when within $5$ pixels from the environment borders, they are repulsed from it.
For more behaviour simulation details, consult the supplementary video.
We collect data for each behaviour in three settings: \texttt{40b} with $40$ agents in a bounded environment, \texttt{30b} with $30$ agents in the same bounded environment and \texttt{40u} with $40$ agents in an unbounded environment.
In each setting, each behaviour is simulated $50$ times, from different initial positions (as defined by the random seed).
We minimise initialisation artefacts that can lead to noisy classification samples by removing the first $250$ steps of each simulation.
% In total, we obtain \num{187500} swarm states (i.e. positions and orientations) per behaviour.
We compute the four feature sets for every swarm state.
As we consider varying swarm sizes, leading to feature set vectors of different sizes (using ``Gomes2013'' and ``Gharbi2023''), we perform a random subsampling schema and only consider $30$ agents' features for the experimental configurations \texttt{40b} and \texttt{40u}.
The dataset and the experimental setup are available at \url{https://kondata.uni-konstanz.de/radar/en/dataset/18h2w28a9d6sm1eq} and \url{https://github.com/alfjesus3/swarm_behaviour_sim}.

The experimental setup for the similarity assessment and classification follows.
The threshold of the swarm mode index feature in the ``Alharthi2022'' is set to \num{0.5}.
The discretisation in the \emph{combined state count} measure is set to low/medium/high with a threshold of $\SI{1e-2}{}$ as suggested in the original paper~\cite{GomChr2013gecco}.
The \emph{sampled average state} measure uses a sampling window of $10$ to allow for a averaging of 1 simulation second.
Across all similarity measures, the behaviour pairs are compared using the same initial conditions, which leads to comparisons of the same behaviour achieving a perfect measure score. 
Therefore, the same-behaviour pairs are omitted from the results.

The hyperparameters for the SOM have been chosen by a grid search.
The map topology is rectangular, with 46 by 46 nodes.
The neighbourhood prototype spreading is Gaussian, and the prototype activation function is Euclidean.
The train-test split for the classification task is $80\%-20\%$.
The training took 180000 SOM update steps with a learning rate of 0.1.
Additionally, each classification sample is a time-series of $5$ sequential simulation steps ($\SI{0.5}{\second}$) and the corresponding numerical class label from 1-6.
For more reliable classification results, the SOM scores are averaged over $3$ independently created and trained models.

\section{Results}
\label{sec:results}
% >>>>> Provide evidence for your claims
% ...

\subsubsection{Similarity}
\label{subsec:results-similarity}

\begin{figure}[tb!]
    \centering
    \begin{subfigure}[t]{0.24\textwidth}
        \centering
        \includegraphics[width=\textwidth]{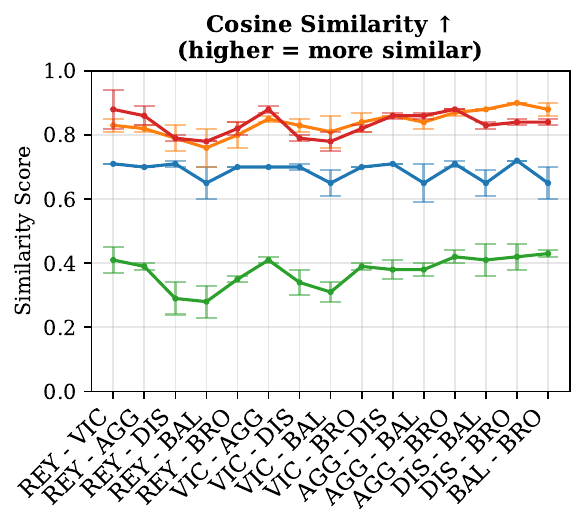}
        \caption{}
        \label{fig:cosine_similarity_avg_behaviours}
    \end{subfigure}
    \hfill
    \begin{subfigure}[t]{0.24\textwidth}
        \centering
        \includegraphics[width=\textwidth]{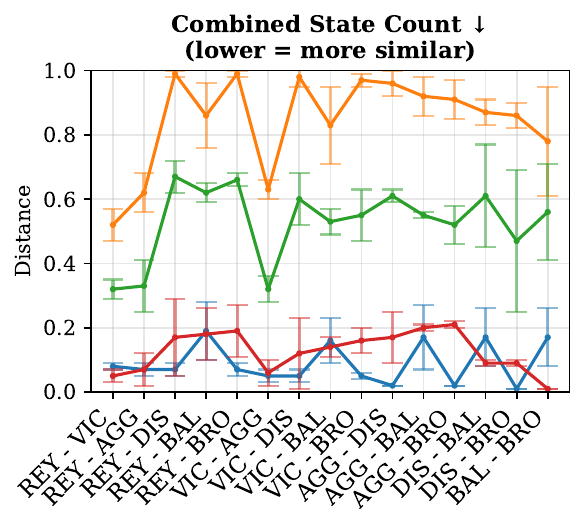}
        \caption{}
        \label{fig:combined_state_count_avg_behaviours}
    \end{subfigure}
    % \vspace{1em} % space between rows
    \begin{subfigure}[t]{0.24\textwidth}
        \centering
        \includegraphics[width=\textwidth]{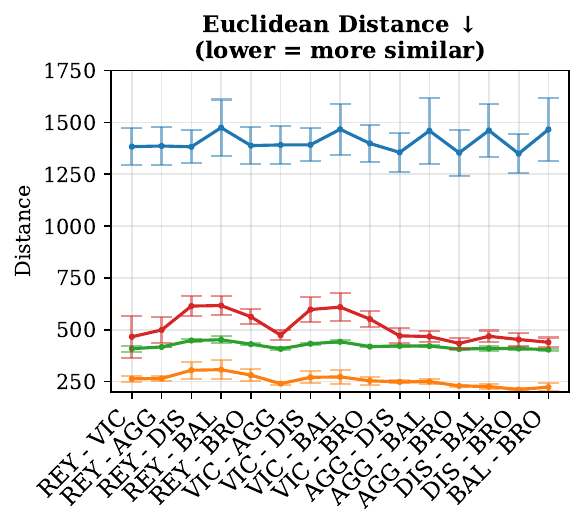}
        \caption{}
        \label{fig:euclidean_dist_avg_behaviours}
    \end{subfigure}
    \hfill
    \begin{subfigure}[t]{0.24\textwidth}
        \centering
        \includegraphics[width=\textwidth]{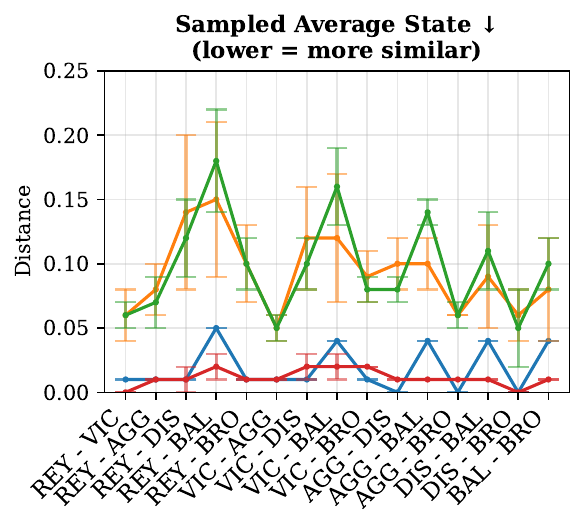}
        \caption{}
        \label{fig:sampled_average_state_avg_behaviours}
    \end{subfigure}
    \includegraphics[width=\textwidth]{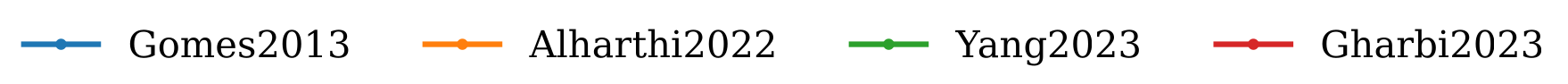}
    \caption{Similarity assessment for all behaviour pairs. Lines represent the mean scores and the standard deviation over $50$ independent simulations.}
    \label{fig:similarity_avg_behaviours}
\end{figure}

Figure~\ref{fig:similarity_avg_behaviours} shows the similarity scores between behaviour pairs.
The \emph{cosine similarity} values (Figure~\ref{fig:cosine_similarity_avg_behaviours}) display the lowest score variability even for behaviours that should, for the most part, be dissimilar, such as flocking and \textit{dispersion}.
In line with the scores spread issue, the absolute similarity values are meaningless by themselves.
% The \textit{Reynolds} and \textit{dispersion} pair (``Alharthi2022'' feature set) with a score of $\num{0.79(4)}$ gives the wrong impression of highly similar behaviours while the \textit{Reynolds} and \textit{Vicsek} pair (``Yang2023'' feature set) with a score of $\num{0.41(4)}$ gives the false impression of dissimilar behaviours.
The ``Alharthi2022'' similarity values display some confusion since the \textit{aggregation} (based on neighbours attraction) and \textit{dispersion} (based on neighbours repulsion) pair has a score of $\num{0.86(0)}$, which is higher than the \textit{Reynolds} and \textit{Vicsek} pair score of $\num{0.83(2)}$ (both flocking behaviours).
I.e., two visually very distinct behaviours (that are often treated as opposites) score higher similarity than two variants of what is arguably the same behaviour.
See the behaviour trajectories in the supplementary video for further visual aid.
% A similar trend happens with the ``Gomes2013'' feature set, where both pairs display the same score of $\num{0.71(0)}$.
% Across all the feature sets, the cosine similarity pairs between \textit{Reynolds}, \textit{Vicsek}, and \textit{aggregation} display consistent scores, highlighting the difficulty of distinguishing these visually similar behaviours (see Figure~\ref{fig:trajs_40b_run0}).

The \emph{combined state count} measure provides the highest score variability, as visually depicted in Figure~\ref{fig:combined_state_count_avg_behaviours}.
Like in the other similarity measures results, the values in absolute terms are meaningless.
The ``Alharthi2022'' feature set is the only set that clearly separates the flocking behaviours (\textit{Reynolds} and \textit{Vicsek}) and the \textit{aggregation} behaviour.
Additionally, this feature set displays dissimilarity pairs near $1$ when there is not much behavioural resemblance, like in flocking--\textit{dispersion} or flocking--\textit{Brownian motion}.
Again, some feature sets suffer from ambiguities, e.g., in the ``Gomes2013'' feature set, the \textit{aggregation}--\textit{dispersion} pair has a score of $\num{0.02(0)}$ while the flocking behaviours pair ($\num{0.08(1)}$).

The \emph{Euclidean distance} measure between the behaviour pairs is outlined in Figure~\ref{fig:euclidean_dist_avg_behaviours}.
In general, it deviates from the distances obtained through the \emph{sampled average state} measure since the two feature sets with the lowest average distances, namely ``Alharthi2022'' and ``Yang2023'', are the feature sets with the highest average distance in the latter measure. 
The ``Gomes2013'' feature set displays the highest distance average and standard deviation (error bars).
%Since the ``Gharbi2023'' feature set uses the shortest neighbour distances, it displays lower distances between the flocking behaviours pair ($\num{466.29(101.44)}$) and between the random walk behaviours pair ($\num{439.93(22.89)}$). 

The \emph{sampled average state} measure provides a distance between the behaviour features (see Figure~\ref{fig:sampled_average_state_avg_behaviours}).
This measure averages over the features of all the agents of the swarm, and subsequently, it averages over all the simulation steps per time window.
Therefore, there is a noticeable difference in the similarity scores between the swarm-level feature sets (``Alharthi2022'' and ``Yang2023'') and the agent-level feature sets (``Gomes2013'' and ``Gharbi2023'') since the latter are represented by a single agent instead of the $N$ agents.
% All the feature sets show a similar distance between \textit{Reynolds}, \textit{Vicsek} and \textit{aggregation} behaviours.

\begin{table}[tb]
\centering
\caption{Classification scores, averaged over three independent runs.}
\label{tab:classification_accuracy_encodings}
\resizebox{\textwidth}{!}{%
\begin{tabular}{@{}ll
S[table-format=1.3(3)]
S[table-format=1.3(3)]
S[table-format=1.3(3)]
S[table-format=1.3(3)]@{}}
\toprule
\multicolumn{2}{c}{\textbf{}} & \multicolumn{4}{c}{\textbf{Feature Sets}} \\
\cmidrule(lr){3-6}
\textbf{Model} & \textbf{Metric} &
{\textbf{Alharthi2022~\cite{AlhAbdHau2022ants}}} &
{\textbf{Gomes2013~\cite{GomChr2013gecco}}} &
{\textbf{Yang2023~\cite{YanSamAru-etal2023IEEEACCESS}}} &
{\textbf{Gharbi2023~\cite{GhaKucGarBir2023icra}}} \\
\midrule

\multirow{2}{*}{\textbf{SOM}} 
& Train accuracy             & 0.489(3) & \bfseries 0.595(5) & 0.500(3) & 0.444(9) \\ 
& Test accuracy              & 0.442(5) & \bfseries 0.494(16) & 0.455(13) & 0.345(5) \\ 

\bottomrule
\end{tabular}%
}
\end{table}

\begin{figure}[tb]
  \centering
  \begin{minipage}[t]{0.36\textwidth}
    \centering
    \includegraphics[width=\linewidth]{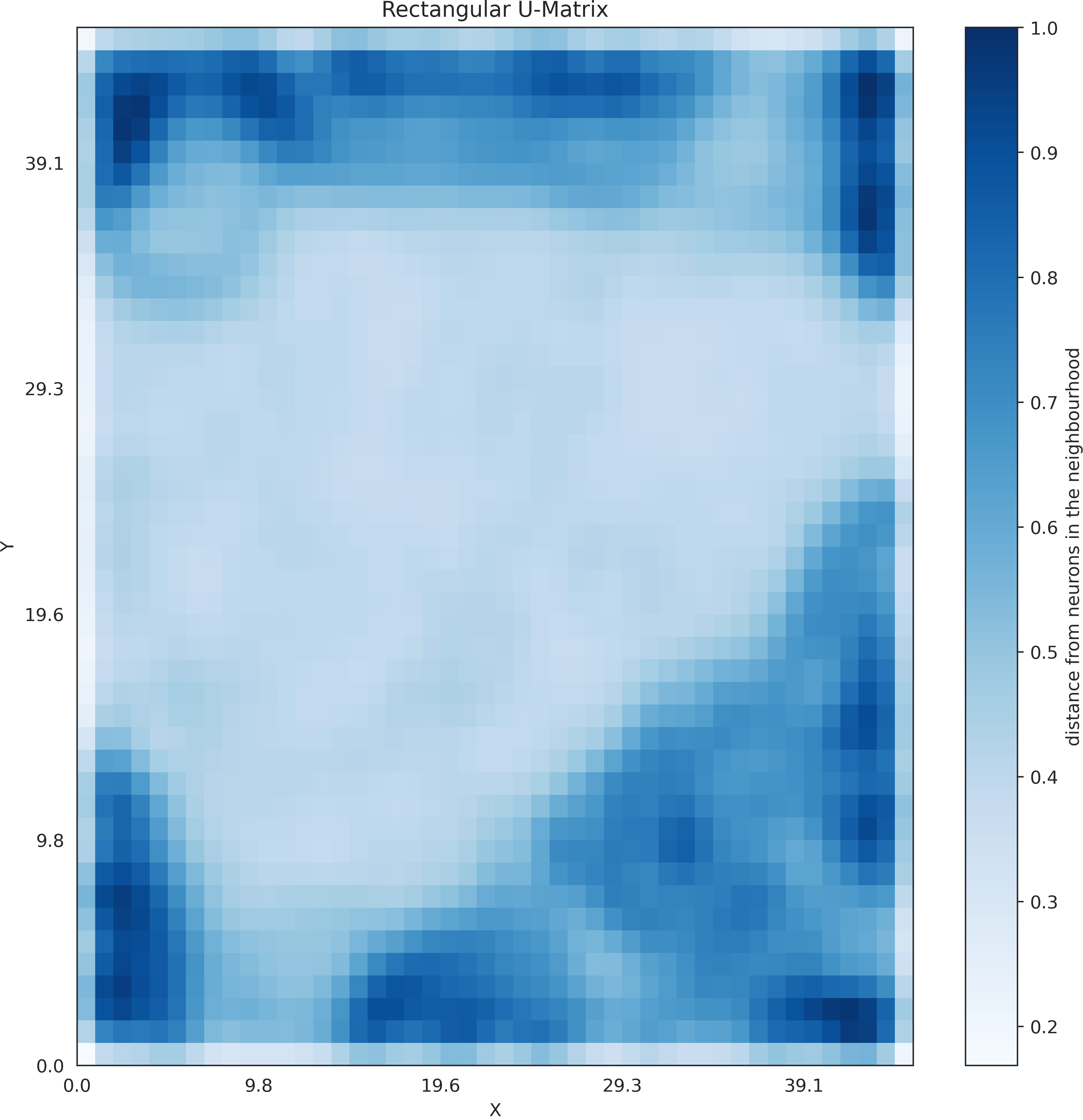}
  \end{minipage}\hfill
  \begin{minipage}[t]{0.6\textwidth}
    \centering
    \includegraphics[width=\linewidth]{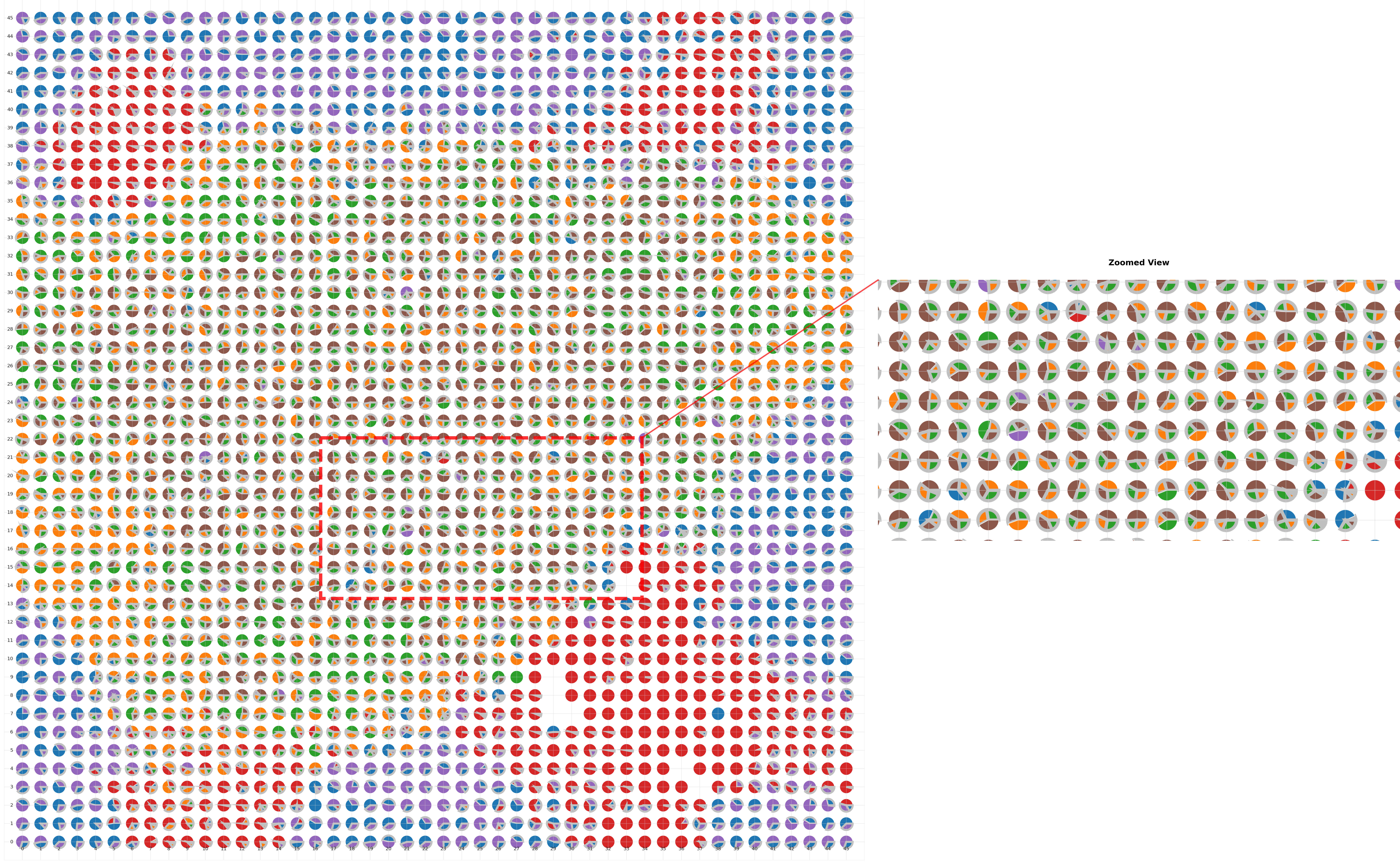}
  \end{minipage}
  \includegraphics[width=\textwidth]{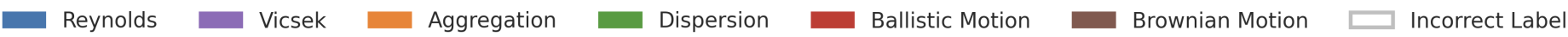}
  \caption{Analysis of the first seed ``Gomes2013'' based SOM. Left: the inter-node Euclidean distances. Right: the training classification.}
  \label{fig:som--gomes2013-tr-class-and-nodes-distance}
\end{figure}

\begin{figure}[tb!]
  \centering
  % Row 1
  \begin{subfigure}[t]{0.32\textwidth}
    \centering
    \includegraphics[width=\linewidth]{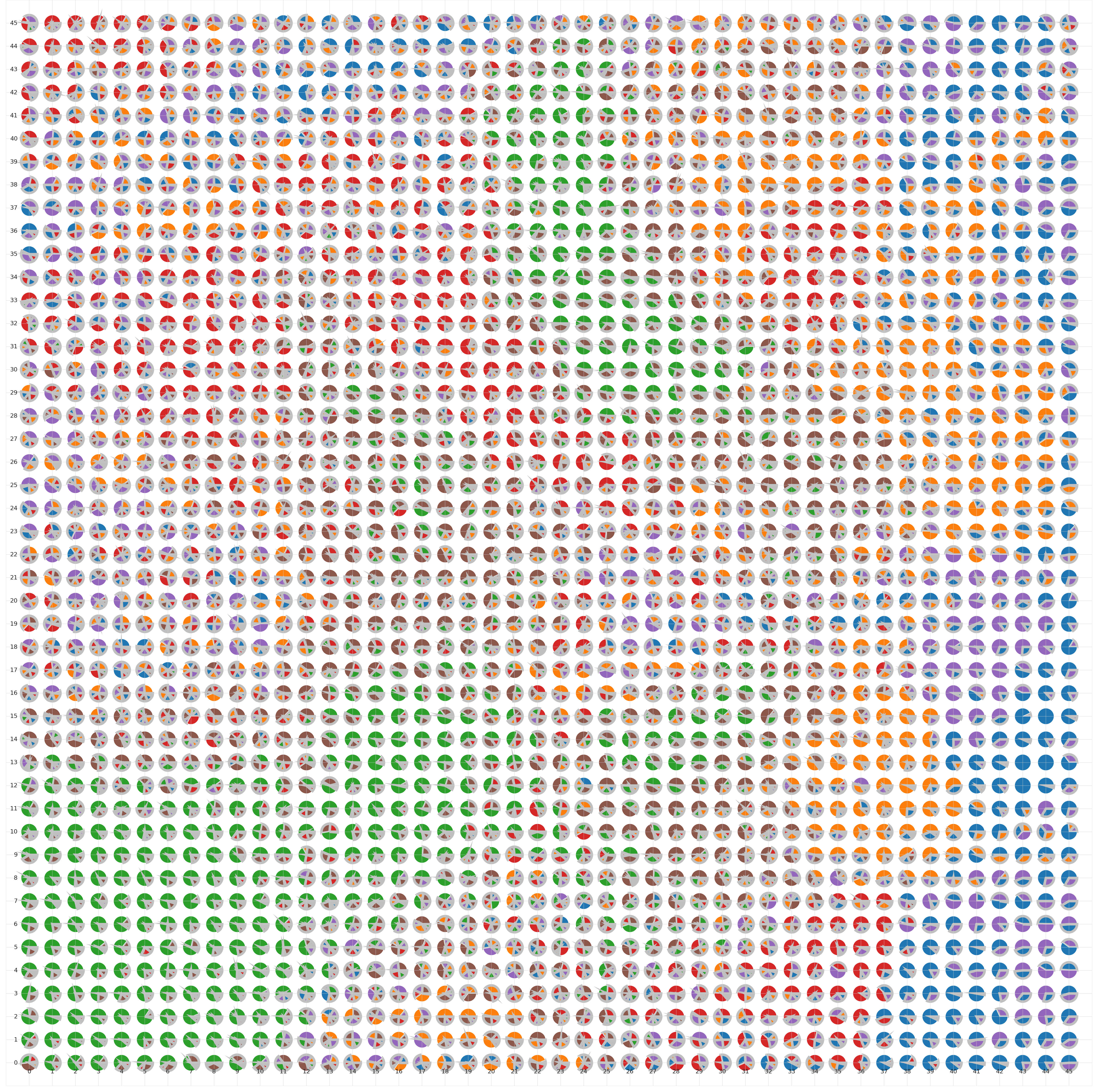}
    \caption{``Alharthi2022''}
    \label{fig:som-Alharthi2022-tr-class}
  \end{subfigure}
  \hfill
  \begin{subfigure}[t]{0.32\textwidth}
    \centering
    \includegraphics[width=\linewidth]{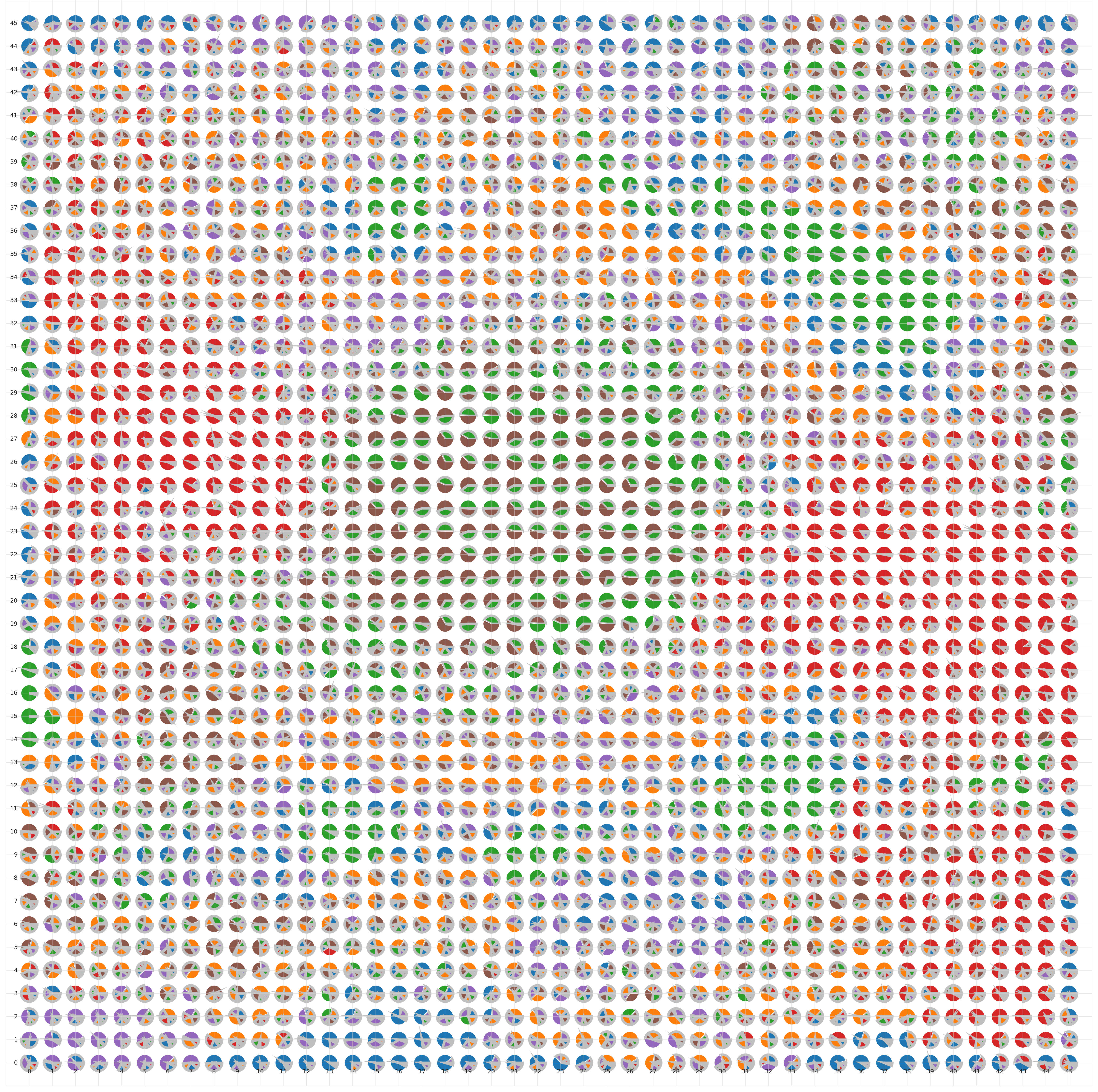}
    \caption{``Yang2023''}
    \label{fig:som-yang2023-tr-class}
  \end{subfigure}
  \hfill
  \begin{subfigure}[t]{0.32\textwidth}
    \centering
    \includegraphics[width=\linewidth]{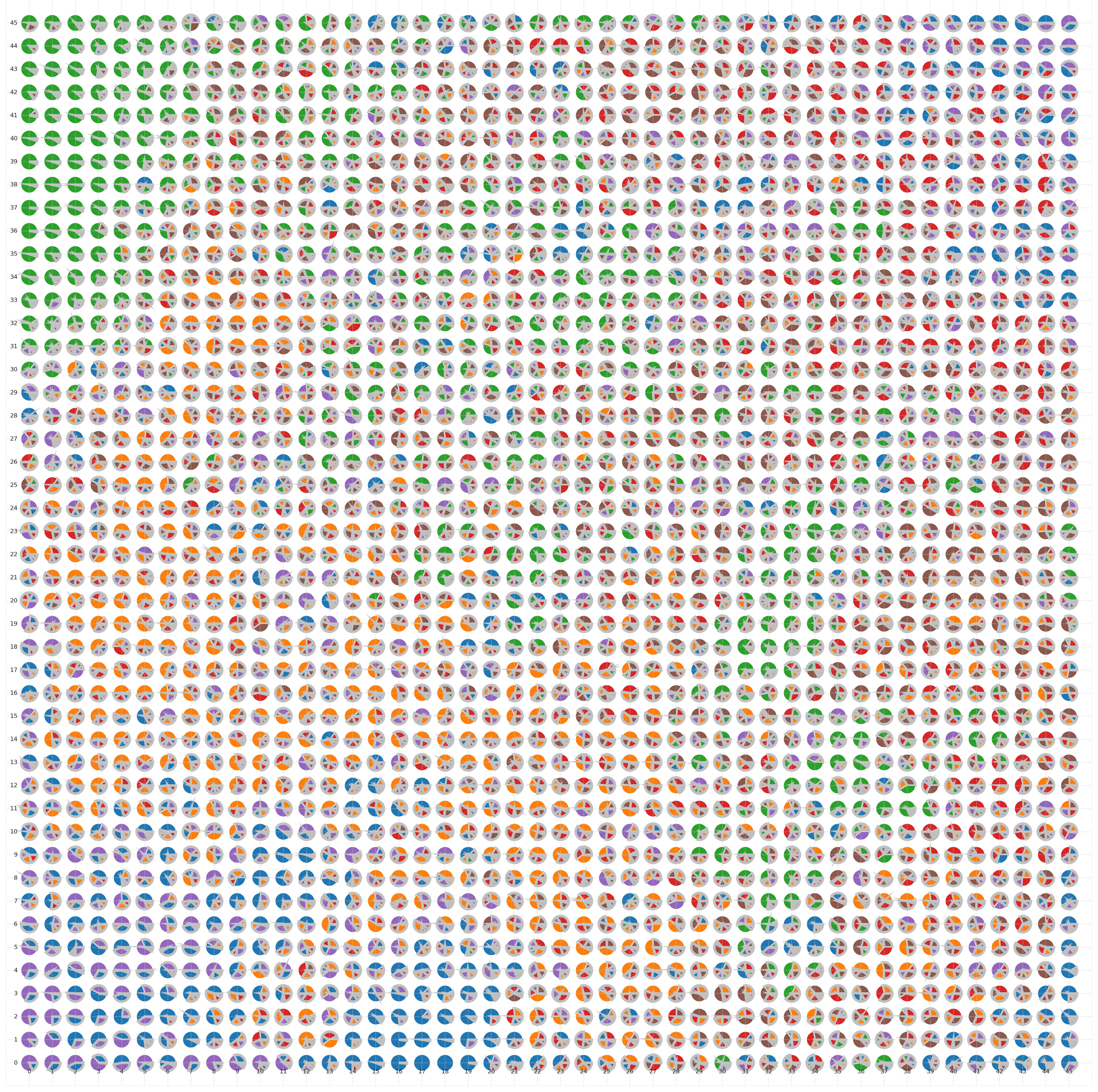}
    \caption{``Gharbi2023''}
    \label{fig:som-gharbi2023-tr-class}
  \end{subfigure}
  % Shared legend
  % \vspace{0.75em}
  \captionsetup{width=\textwidth}
  \includegraphics[width=\textwidth]{figs/som_node_labels_legend.png}
  \caption{Analysis of the first seed SOMs on the training samples. The node colour indicates the true label, and the grey contour indicates incorrect labels. The corresponding inter-node distance maps are available in the supplementary video.}
  \label{fig:som-all-feature-sets}
\end{figure}

\subsubsection{Explainable classification}
\label{subsec:results-classification}
Table~\ref{tab:classification_accuracy_encodings} shows the average scores of the classifiers.
The ``Gomes2013'' (see Figure~\ref{fig:som--gomes2013-tr-class-and-nodes-distance}) feature set achieves the highest accuracy scores as it contains the exact positions and velocities of all agents (i.e., there is no compression of the swarm representation).
However, the accuracy is still relatively low ($\approx\qty{50}{\percent}$ test accuracy).
% The Figure~\ref{fig:som--gomes2013-tr-class-and-nodes-distance} provide further insight into the performance of ``Gomes2013''.
The \textit{ballistic motion} and flocking behaviours are primarily represented in the corners (Figure~\ref{fig:som--gomes2013-tr-class-and-nodes-distance}, right), while the centre area displays confusion in disentangling the representation of \textit{aggregation}, \textit{dispersion}, and \textit{Brownian motion}.
% While there is a (relatively) clear separation between \emph{ballistic motion} and flocking, the two flocking behaviours are often mapped to the same units.
The confusion in the classification of the flocking behaviours is expected, given the discriminatory power of the selected feature sets.
% We did not, however, expect the inability to distinguish between the remaining three behaviours.
Looking at the inter-node distances (Figure~\ref{fig:som--gomes2013-tr-class-and-nodes-distance}, left), we see that the corner regions show higher inter-node distances, while in the centre region the distances are much smaller.
% The inter-node distances are in line with the noticeably higher Euclidean distance results in Figure~\ref{fig:euclidean_dist_avg_behaviours}.
% In the centre region, however, inter-node distances are much smaller.
This is an indicator that the misclassification is not a result of a lack of training data.
The prototypes' closeness implies that any confusion stems from the fact that samples were similar (under the Euclidean distance).

The ``Alharthi2022'' SOM in Figure~\ref{fig:som-Alharthi2022-tr-class} displays a clearer node label separation for the \textit{dispersion} behaviour.
% The misclassification confusion is visible in the many nodes that share predictions from up to $4$ different behaviour classes.
The inter-node distances are more evenly distributed when compared to ``Gomes2013''.
% Additionally, the ``Alharthi2022'' feature set has the lowest quantisation and topological errors (see Table~\ref{tab:classification_accuracy_encodings}).
% Topology-preserving property argument present in "Time Series Prediction with the Self-Organizing Map: A Review" and "Robust Classification with Reject Option Using the Self-Organizing Map"
% While these SOM metrics help the final map preserve the topology of the input samples (such that adjacent patterns are likely to be in adjacent regions of the map), in our case, it does not lead to higher classification accuracies.
The ``Yang2023'' feature set (see Figure~\ref{fig:som-yang2023-tr-class}) based classifier can separate more clearly the nodes with the \textit{ballistic motion} label.
The inter-node distances are comparable, except on the centre cluster that contains the majority of the \textit{Brownian motion} nodes.
These nodes often misclassify the \textit{dispersion} behaviour.
The ``Gharbi2023'' feature set performs the worst in terms of classification accuracy.
This is somewhat expected, as it only considers agents' shortest distances.
This accuracy gap can be, in part, explained by the lack of clear clusters of nodes in the SOM (see Figure~\ref{fig:som-gharbi2023-tr-class}).
Additionally, the low similarity scores variability across all the measures might relate to SOM misclassification.
The poor performance implies that we may need to augment this feature set to help disambiguate the map's topological arrangement.

\section{Discussion and Conclusion}
\label{sec:conclusion}
% >>>>> Discuss the evidence for your claims
In this work, we have investigated the effect of four feature sets from the literature on similarity assessments and classification of collective behaviours.
Our results show that often-used measures, like Euclidean distance or cosine similarity, perform poorly when used to assess the similarity of collective behaviours.
Distance measures particularly designed for swarm robotics (e.g.,~\cite{GomChr2013gecco}) appear to be better suited to compare similarities of collective behaviour.
Additionally, our results show that the most suitable feature set depends heavily on the selected similarity measure and that similarity or distance scores are influenced at least as strongly by the measure as by the compared behaviours.
This highlights the need for future research on the interaction of feature sets and similarity measures.

% >>>>> Remind reader of value and contributions of this paper
% ...
Our classification experiments further highlight the difficulties of finding appropriate feature sets.
The test accuracies of all trained self-organised maps are below \qty{50}{\percent}.
Closer inspection of the SOM prototypes revealed that similar behaviours (e.g., \textit{Reynolds} and \textit{Vicsek} flocking) were often confused, contributing to the lower accuracy.
However, even visually distinct behaviours, such as \textit{dispersion} and \textit{aggregation}, were often matched to the same prototypical units.
This further highlights the fact that selecting features that best represent the collective behaviour is a challenging task. In particular, there is a lack of consistency between the different similarity measures and classification predictors on what features are most useful to discriminate the behaviours.
Our work on self-organised maps provides first insights into this problem, but in the future, we must develop more robust tools to find and choose appropriate features.

\begin{credits}
\subsubsection{\ackname} AFJ and JK acknowledge support from the Carl-Zeiss-Foundation, the DFG through Germany's Excellence Strategy-EXC 2117-422037984 and the Centre for the Advanced Study of Collective Behaviour (CASCB).
JK acknowledges support from the Zukunftskolleg.

\subsubsection{\discintname} The authors declare no competing interests.

\end{credits}

\bibliographystyle{splncs04}
\bibliography{./demiurge-bib-main/definitions,./demiurge-bib-main/author,./demiurge-bib-main/address,./demiurge-bib-main/proceedings,./demiurge-bib-main/journal,./demiurge-bib-main/publisher,./demiurge-bib-main/series,./demiurge-bib-main/institution,./demiurge-bib-main/bibliography,./demiurge-bib-main/additions}

\end{document}